\documentclass[journal]{IEEEtran}

\ifCLASSINFOpdf
\else
\fi
\usepackage{graphicx} 
\usepackage{amsmath}
\usepackage{algorithm}
\usepackage{algpseudocode}
\usepackage{float}
\usepackage{caption}
\usepackage{subfig}
\usepackage{color}
\usepackage{multirow}
\usepackage{cite}

\begin{document}

\title{Tuning Convolutional Spiking Neural Network with Biologically-plausible Reward Propagation
}

\author{Tielin~Zhang, Shuncheng~Jia, Xiang~Cheng and Bo~Xu
\thanks{Tielin~Zhang was with Institute of Automation, Chinese Academy of Sciences (CASIA), Beijing 100190, China, e-mail: tielin.zhang@ia.ac.cn.}
\thanks{Shuncheng~Jia and Xiang~Cheng were with CASIA and School of Artificial Intelligence, University of Chinese Academy of Sciences (UCAS), Beijing 100049, China.}
\thanks{Bo Xu was with CASIA, UCAS and Center for Excellence in Brain Science and Intelligence Technology, Chinese Academy of Sciences, Shanghai 200031, China, e-mail: xubo@ia.ac.cn.}
\thanks{Tielin Zhang and Shuncheng Jia are co-first authors of this paper. The corresponding authors are Tielin Zhang and Bo Xu.}
\thanks{Manuscript received August 13, 2020; accepted May 25, 2021.}}

\markboth{Latex}%
{Zhang \MakeLowercase{\textit{et al.}}}

\maketitle

\begin{abstract}
Spiking Neural Networks (SNNs) contain more biologically realistic structures and biologically-inspired learning principles than those in standard Artificial Neural Networks (ANNs). SNNs are considered the third generation of ANNs, powerful on the robust computation with a low computational cost. The neurons in SNNs are non-differential, containing decayed historical states and generating event-based spikes after their states reaching the firing threshold. These dynamic characteristics of SNNs make it difficult to be directly trained with the standard backpropagation (BP), which is also considered not biologically plausible. In this paper, a Biologically-plausible Reward Propagation (BRP) algorithm is proposed and applied to the SNN architecture with both spiking-convolution (with both 1D and 2D convolutional kernels) and full-connection layers. Unlike the standard BP that propagates error signals from post to presynaptic neurons layer by layer, the BRP propagates target labels instead of errors directly from the output layer to all pre-hidden layers. This effort is more consistent with the top-down reward-guiding learning in cortical columns of the neocortex. Synaptic modifications with only local gradient differences are induced with pseudo-BP that might also be replaced with the Spike-Timing Dependent Plasticity (STDP). The performance of the proposed BRP-SNN is further verified on the spatial (including MNIST and Cifar-10) and temporal (including TIDigits and DvsGesture) tasks, where the SNN using BRP has reached a similar accuracy compared to other state-of-the-art BP-based SNNs and saved 50\% more computational cost than ANNs. We think the introduction of biologically plausible learning rules to the training procedure of biologically realistic SNNs will give us more hints and inspirations toward a better understanding of the biological system's intelligent nature.
\end{abstract}


%
\IEEEpeerreviewmaketitle

\section{Introduction}
\IEEEPARstart{T}{he} rapid development of deep learning (or Deep Neural Network, DNN) breaks many research barriers towards a unified solution by easy and efficient end-to-end network learning. DNNs have replaced many traditional machine learning methods in some specific tasks, and the number of these tasks is still increasing \cite{RN660}. However, during the rapid expansion process of DNNs, many important and challenging problems are exposed accordingly, such as adversarial network attacking \cite{RN740,RN789}, catastrophic forgetting \cite{RN788}, data-hungry \cite{RN790}, lacking causal inference \cite{RN791}, and low transparency \cite{RN792}. Some researchers try to find answers by making efforts on the inner-side research of the DNN itself, such as constructing specific network structures, designing more powerful cost functions, or building finer visualization tools to open the black box of DNNs. These efforts are efficient and have contributed to the further development of DNNs \cite{RN405}. In this paper, we think there are some alternative and easier approaches to achieving these goals, especially on robust and efficient computation, by turning into the biological neural networks and getting inspirations from them \cite{RN699,RN726,RN702,RN732,RN828,RN668,RN96,RN781,RN251}.

The Spiking Neural Network (SNN) is considered the third generation of Artificial Neural Networks (ANNs) \cite{RN679}. The basic information unit transferred between neurons in SNNs is discrete spikes, containing the membrane potential state's precise timing for reaching the firing threshold. This event-type signal contains inner neuronal dynamics and the historically accumulated (and decayed) membrane potential. Spike trains in SNNs, compared to their counterparts, firerates (here we define the firerate as an analog value to describe the propagated information) in ANNs, have opened a new time coordinate for the better representation processing of sequential information. Besides neuronal dynamics, biologically-featured learning principles are other key characteristics of SNNs, describing the modification of synaptic weights by local and global plasticity principles. For local principles, most of them are ``unsupervised'', including but not limited to the Spike-Timing Dependent Plasticity (STDP) \cite{RN691,RN739}, the Short-Term Plasticity (STP) \cite{RN693}, the Long-Term Potentiation (LTP) \cite{RN684,RN764}, the Long-Term Depression (LTD) \cite{RN701}, and the lateral inhibition \cite{RN682,RN795}. For global principles, they are more ``supervised'', with fewer number sizes than local ones, but more related to network functions, e.g., plasticity propagation \cite{RN451}, reward propagation \cite{RN793}, feed-back alignment \cite{zhao2020glsnn}, and target propagation \cite{RN794}.

SNNs are different from structures and functions, such as the echo state machine \cite{RN520}, the liquid state machine \cite{RN529}, the feed-forward architecture with biological neurons \cite{RN745,RN732}, and some task-related structures \cite{RN704,RN756,RN501}. Some tuning methods of SNNs are BP-based (e.g., ReSuMe \cite{RN491}, SpikeProp \cite{RN752}, and ANNs trained with BP first and then converted into SNNs \cite{RN484}) or BP-related (BP through time \cite{RN648,RN643}, SuperSpike \cite{RN424}, and STDP-type BP \cite{RN641}). There are still some efforts to train SNNs with biologically plausible plasticity principles (e.g., STDP or STP-based learning \cite{RN691,RN470,RN468,RN696,RN694,RN761,RN693}, equilibrium learning \cite{RN773,RN760,RN562}, multi-rule-integrative learning \cite{RN711,RN758,RN256}, and curiosity-based learning \cite{RN762}).

This paper focuses more on training SNNs with biologically plausible learning principles instead of directly tuning them with BPs to get closer to understanding the brain first and then achieving human-level artificial intelligence. Hence, the SNN with spiking-convolution layers (containing both 1D and 2D spiking-convolution kernels) and full-connection layers (containing neuronal dynamics) is constructed and tuned with the proposed Biologically-plausible Reward Propagation (BRP), named BRP-SNN. The BRP-SNN focuses on describing the neuron-level dynamic computation of membrane potential and the network-level rewiring of global synapses. Unlike other tuning methods that usually focus more on local plasticity principles (e.g., the STDP and the STP) or mixed local and global principles, we use the global BRP only for simplicity and clarity. After the global reward propagation, synaptic modifications are further induced with only local gradient differences that can also be replaced with the STDP and the differential Hebb's principle.

The main contributions of this paper include the following:
\begin{itemize}
\item We construct a multi-layer SNN architecture containing LIF neurons and a network with efficient spiking-convolution and spiking-pooling layers.
\item We propose a new Biologically-plausible Reward propagation (BRP) for the efficient learning of SNNs. Unlike other BP-based tuning algorithms, the BRP tunes the hidden neurons globally with only labels from the output layer instead of layer-by-layer error backpropagation. This new method has a much lower computational cost for reaching the same performance. Besides, the reward information is more biologically plausible than other error-based signals in the biological system. We think this is an alternative effort to achieve biologically efficient learning by tuning biologically realistic SNNs with biologically plausible plasticity principles.
\item We use both spatial (including MNIST and Cifar-10) and temporal (including TIDigits and DvsGesture) datasets to test the proposed algorithm's performance on SNNs. The BRP-SNN achieves high performances (accuracy on test sets, 99.01\% for the MNIST, 57.08\% for the Cifar-10, 94.86\% for the TIDigits, 80.90\% for the DvsGesture), compared with that by other state-of-the-art (SOTA) SNNs tuned with pure biologically-plausible principles. Furthermore, the BRP-SNN also shows power on a lower computation cost (saving more than 50\% neuron computation).
\end{itemize}

The remainder of the paper is organized as follows. Section \ref{Background} gives a brief introduction of the information processing in biological neurons and networks. Section \ref{Method} describes the multi-layer SNN with spiking-convolution and spiking-pooling layers, which is then tuned by three types of target propagation methods (including BRP). Section \ref{Experiments} introduces the comparable convergence, lower computational cost, and better (or compatible) performance of BRP-SNN compared to other SOTA algorithms tuned with biologically plausible (or BP-based) learning methods. Finally, some conclusions are given in Section \ref{Conclusion}.

\section {Background of biologically-plausible information processing}\label{Background}

\subsection{Dynamic spiking neurons}

Dynamic neurons in SNNs are very different from their counterpart activation functions in DNNs. The neurons communicate to each other in different layers of SNNs with discretely event-based spikes but not continuous firerates in DNNs. DNNs also contain various ``neurons'' (activation functions), such as the Rectified Linear Unit (ReLU), the Sigmoid function, and the Tanh function. However, they describe only the spatial non-linear mapping between neurons' input and output signals instead of temporally wiring them with neuronal dynamics.

\begin{figure}[htp]
\centering
\includegraphics[width=8.8cm]{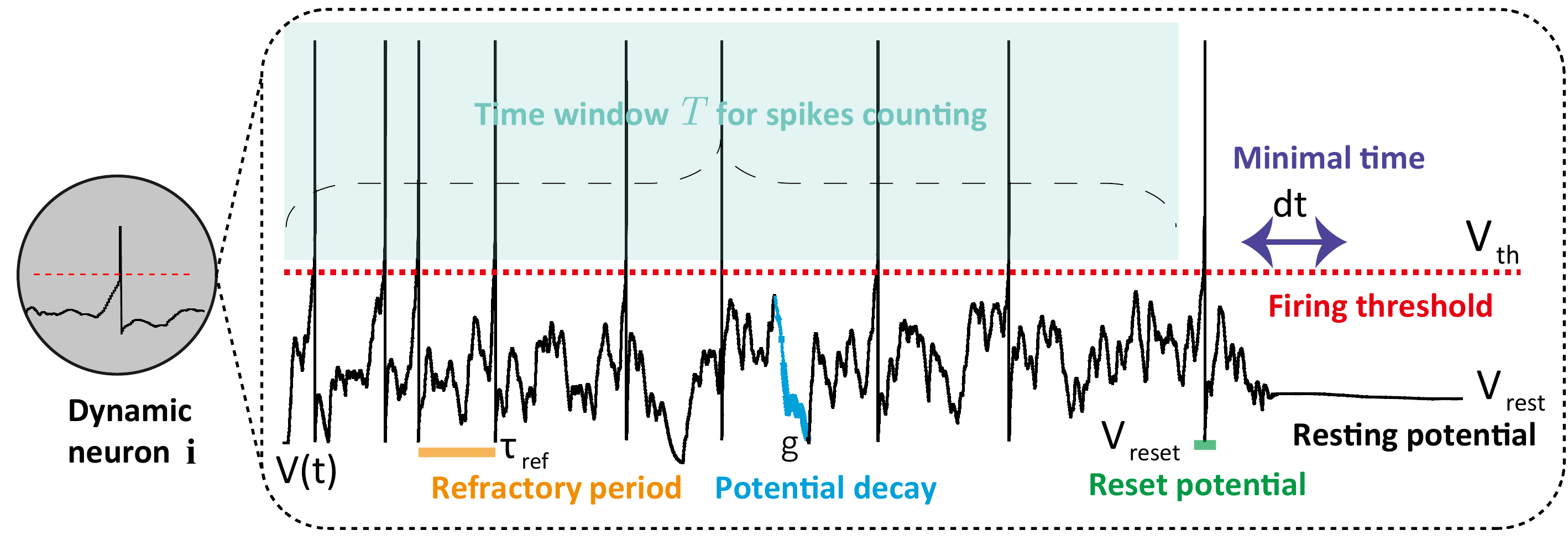}
\caption{An example of dynamic LIF neurons in SNNs.}
\label{fig_lif}
\end{figure}

Basic computational units in SNNs are different types of dynamic neurons \cite{RN745}, for example, the Hodgkin-Huxley (H-H) neuron, the Leaky Integrated-and-Fire (LIF) neuron, the Izhikevich neuron, and the spike response neuron (SRM). The LIF neuron describes the dynamics of membrane potentials (shown in Fig. \ref{fig_lif}). The typical LIF neuron dynamically updates its membrane potential $V_i(t)$ in a long-range time domain, according to:

\begin{equation}
\left\{\begin{array}{l}
C\frac{dV_i(t)}{dt}=g\left(V_i(t)-V_{rest}\right)+\sum_{j=1}^NW_{i,j}X_j(t)\\
\begin{matrix}
V_i(t)=V_{reset} & if(V_i(t)=V_{th}, t-t_{spike}>\tau_{ref})
\end{matrix}
\end{array}\right.\text{,}
\end{equation}

where $\tau_{ref}$ is refractory period, $g$ is synaptic conductivity, $V_{reset}$ is reset value after $V_i(t)$ reaches firing threshold $V_{th}$. The resting membrane potential $V_{rest}$ is set as an attractor of $V_i(t)$ especially when no stimulus input is given. $C$ is membrane capacitance. $dt$ is minimal time step for $V_i(t)$ update (usually set as 0.1-1 $ms$). $j$ is neuron index of presynaptic neuron. $i$ is current neuron index. $N$ is number of neurons in the current layer. $W_{i,j}$ is synaptic weight between the presynaptic neuron $j$ and the current neuron $i$. $t_{spike}$ is specific spike time of the neuron $i$. $X_j(t)$ is neuronal input from the presynaptic neuron $j$. All these neuronal dynamics will proceed within a time window $T$.

\subsection {The biological plausibility of the BRP}

The computation in the biological network is asynchronous, containing obviously multi-clocks at different time scales, from synapses, neurons to networks \cite{xing2020fronti,RN561}. The biological brain handles this multi-scale information processing challenge with event-based spikes containing spatial and temporal signals. These complex dynamics of spikes also make the conventional BP-based algorithms difficult on the network learning. 

\begin{figure}[htp]
\centering
\includegraphics[width=8.5cm]{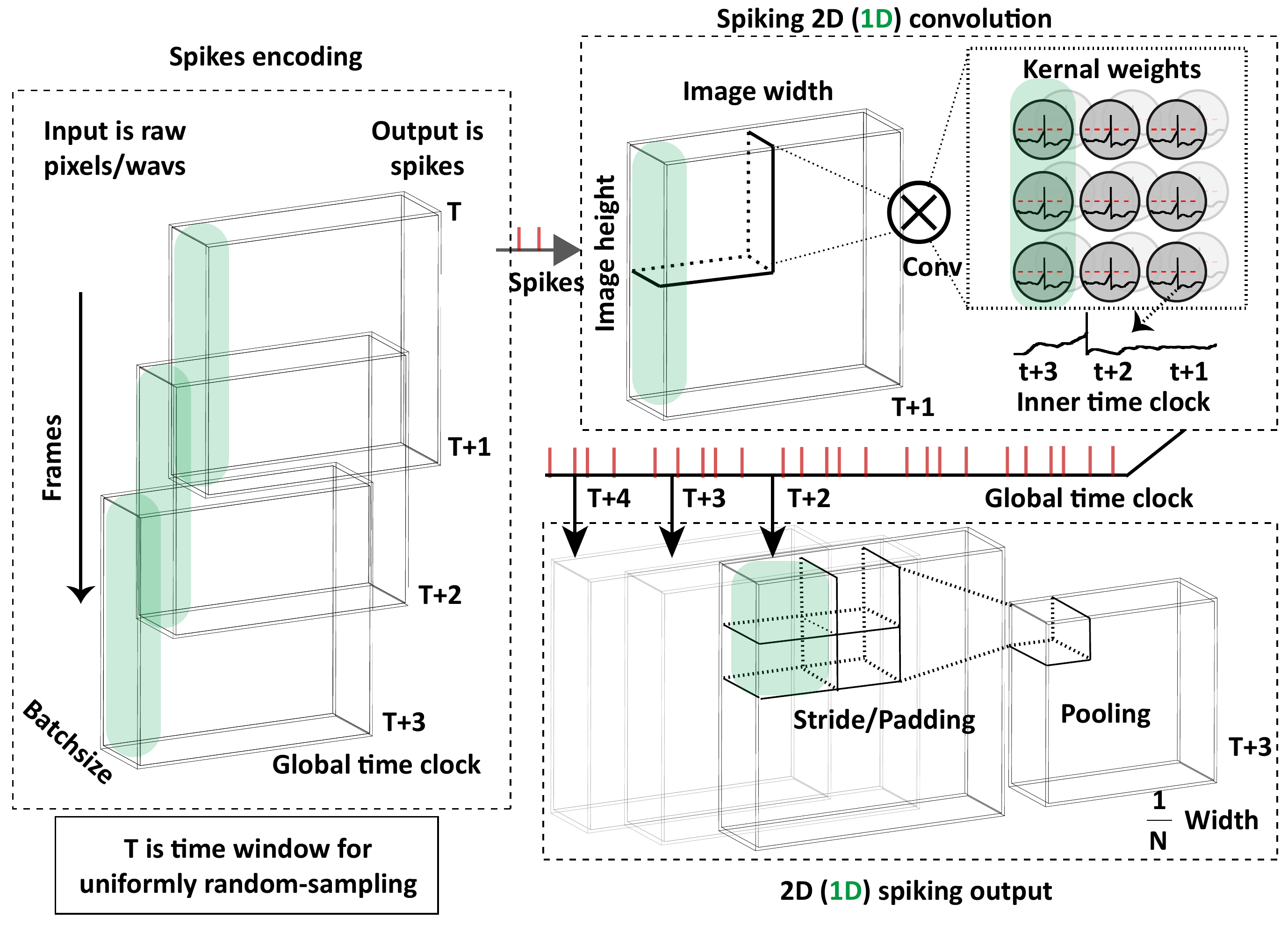}
\caption{The multi-scale information processing encoded with multi-clock spikes, during the 2D (or 1D) spiking-convolution and spiking-pooling layers.}
\label{fig_conv}
\end{figure}

Fig. \ref{fig_conv} shows an example of spikes information processing at convolution and pooling layers, where multi-clock spikes (including local time clock inner neurons and global ones between layers) are integrated for the generally complex information encoding.

The biological system will release dopamines from neurons in the basal ganglia directly to other brain circuits (e.g., rerouting by thalamus) to change the synaptic states. This procedure has inspired us to propose the BRP algorithm, which is efficient and different from the BP in ANNs\cite{RN694}.

\section{Method}\label{Method}

Here we add a $Spike$ flag into standard LIF neurons in order to slow down the update speed of $V_i(t)$ instead of absolutely blocking it during the refractory period, shown as follows:

\begin{equation}
\left\{\begin{array}{l}
\begin{matrix}
C\frac{dV_i(t)}{dt} = &g(V_i(t)-V_{rest})(1-Spike) \\
& +\sum_{j=1}^NW_{i,j}X_j(t)
\end{matrix}\\
\begin{matrix}
V_i(t)=V_{reset}, Spike=1& if(V_i(t)=V_{th})
\end{matrix}\\
\begin{matrix}
Spike=1 & if(t-t_{spike}<\tau_{ref}, t\in(1,T))
\end{matrix}
\end{array}\right.\text{,}
\end{equation}

where $V_i(t)$ is the integration of its historical membrane potential $V_i(t-k)$ ($k\in{1,2,3,..}$) with decay $g^k$ and its current input stimulus $X_j(t)$. The $g$ (i.e., conductivity) is the decay factor of $V_i(t)$ which is usually designed as a hyper-parameter with a value smaller than 1 nS (nanosecond). $Spike$ will be generated if $V_i(t)$ reached the firing threshold $V_{th}$, and at the same time, $V_i(t)$ will also be reset as a predefined membrane potential $V_{reset}$. Furthermore, a parameter of refractory time period $\tau_{ref}$ is given, during which $V_i(t)$ will have a much lower update speed. $V_{rest}$ can also be considered as one attractor of $V_i(t)$ especially when no input $X_i(t)$ is given and no $Spike$ is generated, where the membrane potential $V_i(t)$ will be dynamically decayed into $V_{rest}$. The inner iterative time window $T$ for calculating neuronal dynamics is in a range of 10 to 100 milliseconds.

\subsection{Spiking-convolution layer with 1D and 2D kernels}

The 1D and 2D convolutions have been proved efficient on temporal and spatial information processing, respectively. The reusability of convolutional kernels (i.e., the sharing of synaptic weights) also contributes to network learning towards the anti-overfitting characteristic of SNNs to some extent.

As shown in Fig. \ref{fig_conv}, each layer's signal is represented as spike trains. For simplicity, the time clock $t$ that inside the LIF neuron for the membrane potential update is designed as the same as the outside neuron clock $T$ used for the signal propagation from previous to post layers. This configuration means the time range of information propagation within and outside of neurons is all $T$ for computation ease. The convolutional layer 2D (or 1D) spiking input with height and weight dimensions (or with only the height dimension for the 1D input, i.e., green bars in Fig. \ref{fig_conv}), and then makes the convolution with 2D (or 1D) kernels. These neurons in kernels are designed as dynamic neurons containing spikes that encode event-based signals with a learning time $t$ going by (until $T$). Then the spikes after convolutions further proceed with pooling for the dimension reduction.

\begin{figure}[htp]
\centering
\includegraphics[width=8.5cm]{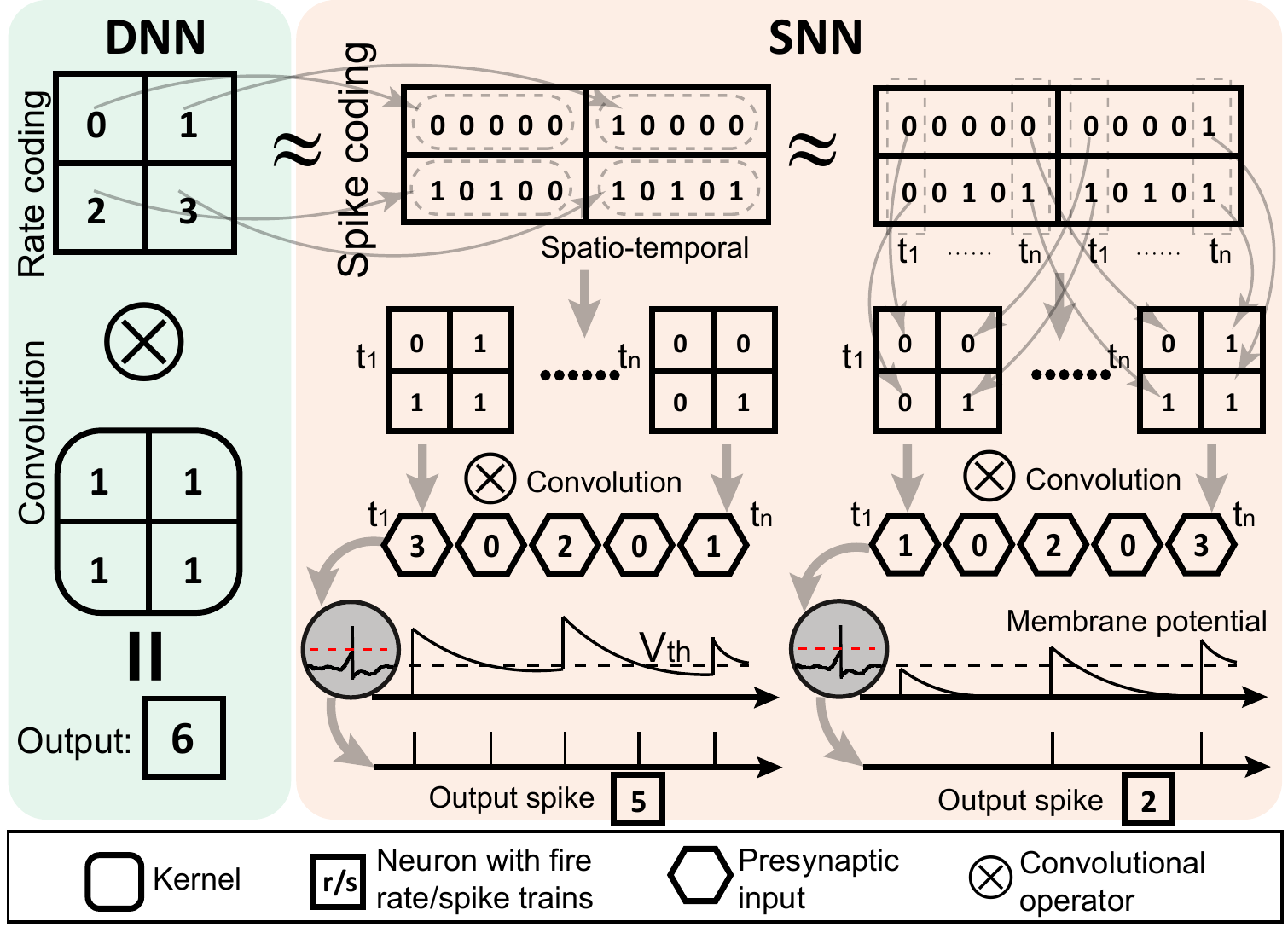}
\caption{Schematic diagram depicting different procedures of the standard convolution in DNNs and the spiking convolution in SNNs, where the spiking neuron of SNNs can generate different outputs given input signals with same firerates (but with different sequence orders).}
\label{fig_compare_convs}
\end{figure}

The convolutional processing in SNNs is different from that in traditional DNNs. For a clearer description of their differences, a schematic diagram of two types of convolutions is described and shown in Fig. \ref{fig_compare_convs}. In SNNs, the decay of membrane potentials will make LIF neurons more difficult to fire. On the contrary, membrane potentials' historical positive integrations will make LIF neurons generate relatively more output spikes. These two dynamical parts in the LIF neuron will contribute to the temporal information processing of SNNs at the micro-scale neuron level. On the contrary, ANNs generate the same firerates as outputs by ignoring the sequential differences of input signals. This sequence-related dynamics of LIF neurons will contribute to the temporal information representation and learning in SNNs.

\begin{figure*}[htp]
\centering
\includegraphics[width=15cm]{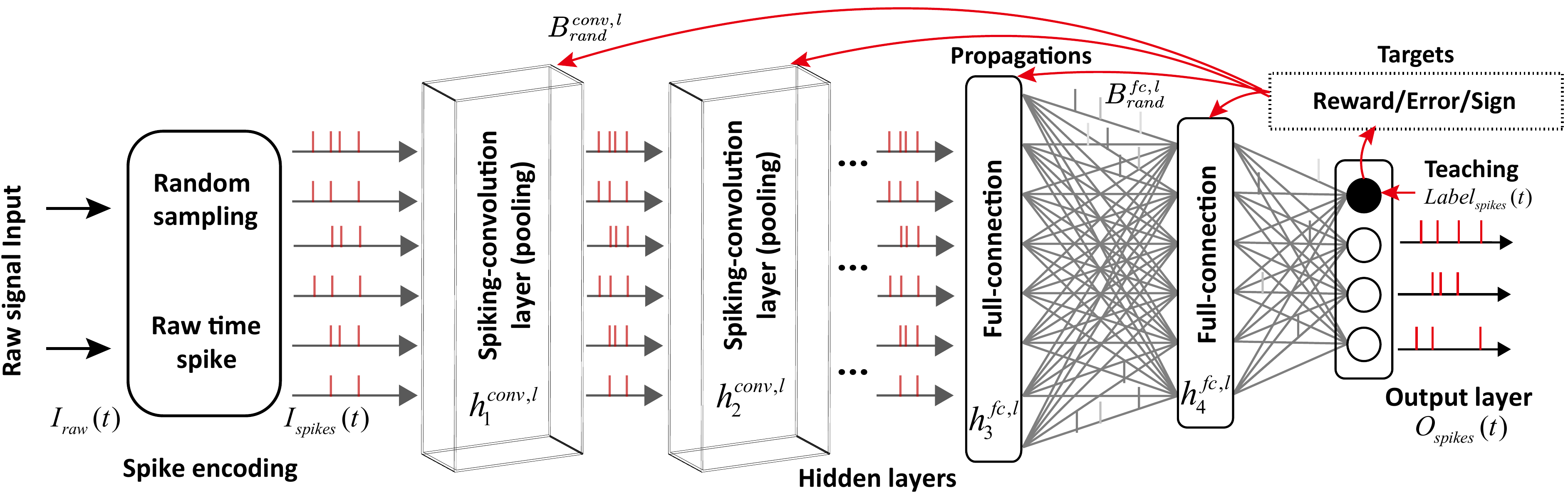}
\caption{The architecture of the BRP-SNN.}
\label{fig_architecture_SNN}
\end{figure*}

\subsection{The BRP-SNN architecture}

The whole BRP-SNN architecture is shown in Fig. \ref{fig_architecture_SNN}, where input spike trains are generated and encoded from the raw signal input, which can be the 1D auditory signal or the 2D image signal. For 1D and 2D signals without spikes (e.g., TIDigits and Cifar-10), the raw input signal $I_{raw}(t)$ is randomly sampled as sequential spike trains $I_{spikes}(t)$ first, and then propagated into the following convolutional, pooling, and full-connection layers. For signals with natural spikes, such as 3D dynamic-vision-sensor (DVS) video signals (e.g., DvsGesture), the raw signal has already been well encoded as 0 or 1 event-type symbols, which can be considered as spikes directly. Hence the additional spike encoding is not necessary anymore. The spike generator is shown as follows:

\begin{equation}
I_{spikes}(t)=\sum_{i=1}^T\delta(t-t_i)\left(I_{raw}(t)<I_{rd}(\alpha)\right)
\label{equa_spike_generator}\text{,}
\end{equation}

where $I_{rd}$ is a $0-1$ random variable generator with a decay factor $\alpha$, and $I_{rd}(\alpha)=I_{rd}*\alpha$. $T$ is the global network clock, which is the same as the inner clock of LIF neurons.

The spiking-convolution and pooling layers are constructed for spatial feature detection, and the feedforward full-connection layers are constructed for the next-step classification. Spike trains of different hidden layers in SNNs can be summed as firerates first after the whole time window $T$, and then combine propagated label-reward signals to modify local synaptic weights.

\subsection{Tuning SNNs with the global BRP}

Feed-forward and feed-back propagations are usually interleaved together for the convergence learning of SNNs. The feed-back propagation is also described as a top-down refinement of the network structure (e.g., synaptic weights) with directly error-to-synapse modification (e.g., revised or pseudo-BPs) \cite{RN726,ieee8325230}. However, these BPs are considered not biologically plausible for their layer-by-layer error backpropagation.

In the biological system, the signal can backpropagate within inner neurons or only within neighborhood layers \cite{RN508,RN432,RN367,RN726,RN678}. Long-range propagation describes the reward propagation from neurons in the brain region for higher cognitive functions directly to target neurons in that for primary cognitive functions \cite{RN726}. The idea of a direct random target propagation algorithm \cite{RN794} fits these biological constraints. Hence, we select it as the underlying bone architecture of the BRP-SNN and further refine it with neuron dynamics, spiking convolutions, and reward propagations.

The biologically plausible reward propagation BRP is constructed according to:

\begin{equation}
TP_{BRP}=Label_{spikes}(t)=\bar{y}(t)\text{,}
\end{equation}

describing a spike train to represent target labels (i.e., the one-shot $Label_{spikes}(t)$ during $T$) as the reward to be propagated to all of pre-hidden layers directly. The $TP$ is a short name of ``Target Propagation'', which means giving the target goal from the output layer to all of the previous layers directly. Different from $TP_{BRP}$, another candidate $TP_{err}$ is also constructed, shown as follows:

\begin{equation}
TP_{err}=y(t)-\bar{y}(t)\text{,}
\label{equa_TP_err}
\end{equation}

where $TP_{err}$ describes the difference between output signals and teaching signals. $\bar{y}(t)$ is the mean firerate of teaching signals (e.g., labels). $y(t)$ is the mean summation of output spike trains from SNNs ($O_{spikes}(t)$), shown as follows:

\begin{equation}
\left\{\begin{array}{l}
y(t)=\frac{1}{T}\sum_{t=1}^TO_{spikes}(t) \\
\bar{y}(t) = \frac{1}{T}\sum_{t=1}^TLabel_{spikes}(t)
\end{array}\right.\text{,}
\label{equa_TP_y}
\end{equation}

where spikes in the time window $T$ are summed together as the firerate $y(t)$ only at the end of $T$. $TP_{sign}$ is the third type of TPs, which describes the sign of positive and negative errors, shown as follows:

\begin{equation}
TP_{sign}=sign(y(t)-\bar{y}(t))\text{,}
\label{equa_TP_sign}
\end{equation}

where the detailed error signal is not necessary, and only the sign of errors is used for the propagation. This effort propagates only positive (``+'') and negative (``-'') signals to hidden layers. However, this sign's calculation contains the premise that detailed errors have to be calculated first and then quantitatively normalized as signs of errors. Hence it still needs both $y(t)$ and $\bar{y}(t)$, the same with $TP_{err}$, instead of the pure reward definition related only with the functional goal (where $y(t)$ is not necessary) of the network such like $TP_{BRP}$. This idea is not new, and some previous work has described the sign instead of error propagation to the network. NormAD \cite{ijcnn7280618} is a good example in them, which can be seen as an efficient learning rule by the sign propagation, especially in shallow SNNs.

Different types of TPs will directly propagate to all hidden layers, including spiking-convolution layers and full-connection layers. The additional $B_{rand}$ will be given as a randomly generated matrix during the network initialization, which will play roles in the different-dimensional matrix conversion from the output layer to all hidden layers. Notice that the global propagation can only propagate signals and affect neuron states (e.g., membrane potentials) instead of synaptic weights. Hence, the synaptic modification for the learned knowledge will be further processed in the next-step procedure of local synaptic weight consolidation.

\subsection{The local synaptic weight consolidation}

The BRP will propagate back directly into different hidden layers of SNNs at each time $t$. This effort will influence the current network states, and at the same time, give SNNs a desired target state for the next-step learning at the time $t+1$. Synaptic weights in SNNs can be updated based on the difference of locally propagated neural states $B_{rand}^{conv,l}*TP_{BRP}(t)$ and current neural states $h^{conv,l}(t)$. The modification of synaptic weights $\Delta W_{i,j}^{conv,l}$ in spiking-convolution layers are updated as follows:

\begin{equation}
\Delta W_{i,j}^{conv,l}=-\eta^{conv}(B_{rand}^{conv,l}*TP_{BRP}(t)-h^{conv,l}(t))\text{,}
\label{equa_W_conv}
\end{equation}

where $B_{rand}^{conv,l}$ is a randomly generated matrix for the hidden layer $l$ before network learning and will not be further updated. $\Delta W_{i,j}^{conv,l}$ is the modification of synaptic weights in the convolutional layer $l$ from the presynaptic neuron $j$ to the post synaptic neuron $i$. $\eta^{conv}$ is the learning rate. $h^{conv,l}(t)$ is the neuron state (the same as the membrane potential $V$) in the current layer $l$. Similar to that, the modification of $\Delta W_{i,j}^{fc,l}$ in the full-connection layer is updated as follows:

\begin{equation}
\Delta W_{i,j}^{fc,l}=-\eta^{fc}(B_{rand}^{fc,l}*TP_{BRP}(t)-h^{fc,l}(t))\text{,}
\label{equa_W_fc}
\end{equation}

where $B_{rand}^{fc,l}$ is also a randomly generated matrix for the hidden layer $l$ before network learning and will not be further updated during learning. $\Delta W_{i,j}^{fc,l}$ is the synaptic weight in full-connection layer $l$ from the presynaptic neuron $j$ to the post synaptic neuron $i$. $\eta^{fc}$ is learning rate, and $h^{fc,l}(t)$ is membrane potential state in the current layer $l$.

\subsection{The pseudo gradient approximation}

The dynamic LIF neuron in SNNs is non-differential, which is difficult to propagate the gradient for further calculating the gradient from the global BRP to the local weight consolidation. Traditional local tuning of SNNs uses STDP or differential Hebb's principle for the locally synaptic modification \cite{RN696,RN691,RN761}. Unlike that, here, we give an additional pseudo gradient approximation of LIF neurons to tune SNNs with the Pytorch architecture \cite{paszke2017automatic}. The pseudo gradient converts the non-differential procedure of firing neurons (where the membrane potential $V_i(t)$ reaching the firing threshold $V_{th}$ is then reset as $V_{reset}$) as a specific pseudo gradient. The pseudo gradient approximation is calculated as follows:

\begin{equation}
\begin{matrix}
\Delta V_i(t)=V_i(t+1)-V_i(t)=1 & if(V_i(t)=V_{th})
\end{matrix}\text{,}
\label{equa_approx_gradient}
\end{equation}

where $\Delta V_i(t)$ is set as a finite number (here is 1 for simplicity) during the propagation, in order to bypass the non-differential problem of $V_i(t)$ during the traditional procedure of gradient propagations, where the original $\Delta V_i(t)$ is actually infinite when $V_i(t)=V_{th}$.

\subsection{The algorithm complexity analysis}

One advantage of using biologically plausible algorithms to tune K-layer SNNs is the lower algorithm complexity. As shown in Table \ref{tab_complexity}, the pure STDP updates synaptic weights directly during the feedforward information propagation. Hence the algorithm complexity of STDPs is $O(nK+mK)$, where $n$ and $m$ are computational costs of one-step feedforward and one-step local-plasticity propagation, respectively.

\begin{table}[htb]
\caption{Algorithm complexity for tuning K-layer SNNs}
\centering
\begin{tabular}{|c|c|}
\hline
Strategy & Complexity \\
\hline
Pseudo BP &$O(nK+(m+mK)K)$ \\
\hline
BRP &$O(nK+(m+1)K)$ \\
\hline
Pure STDP &$O(nK+mK)$ \\
\hline
\end{tabular}
\label{tab_complexity}
\end{table}

The BRP finishes the feedforward procedure with $O(nK)$, and then parallelly propagates reward from the output layer to all hidden layers with an additional one-step matrix conversion ($B_{rand}^{conv,l}$ and $B_{rand}^{fc,l}$). Hence, the algorithm complexity of the BRP is $O(nK+(m+1)K)$. For pseudo-BPs, besides the feedforward cost with $O(nK)$, it also contains multi-step back propagations layer by layer with differential chain rule, costing $O(nK+(m+mK)K)$.

\subsection{The BRP-SNN learning procedure}

The detailed learning procedure of the BRP-SNN algorithm is shown in Algorithm \ref{algorithm1}.

\begin{algorithm}
\caption{The BRP-SNN learning procedure.}
\label{algorithm1}
\begin{algorithmic}
\State 1. Start network initialization: Pack raw datasets as input with or without spikes generation by random sampling; Initialize all of hyper parameters related to dynamic neurons $C$, $g$, $dt$, $V_{rest}$, $W_{i,j}$, $V_{reset}$, $\tau_{ref}$, and other related parameters of networks, e.g., kernel size, kernel numbers, number of hidden layers and neurons, learning rates $\eta^{conv}$ and $\eta^{fc}$, propagation matrixes of $B_{rand}^{conv,l}$ and $B_{rand}^{fc,l}$ in each layer $l$, and time window $T$;
\State 2. Start the training procedure:
\State (1). Load training samples;
\State (2). Feedforward information propagation of $I_{spikes}$;
\State (3). Iteratively increase the network clock $t$ and output $O_{spikes}$ until $t$ reaches the time window $T$;
\State (4). Calculate the firerate of network $y(t)$ until $T$;
\State (5). Apply BRP, $TP_{err}$ or $TP_{sign}$ on spiking-convolution and full-connection layers for different tasks, respectively;
\State (6). Local synaptic modifications with pseudo gradients;
\State (7). Iteratively train BRP-SNN from Step (2)-(7) until convergence; Save the tuned synaptic weights in all layers.
\State 3. Start the testing procedure:
\State (1). Load test samples;
\State (2). Test the performance of BRP-SNNs with feedforward propagation based on saved synaptic weights in each layer.
\State (3). Output the neuron index with a maximum firerate $y(t)$ in time window $T$ as the target class; Calculate the accuracy in the whole test dataset without cross-validation.
\State 4. Finish SNN learning.
\end{algorithmic}
\end{algorithm}

\subsection {The analysis of why BRP works}

The proposed BRP is innovative and will be further verified to finish the following spatial and temporal classification tasks in Section \ref{Experiments}. The BRP is very different from the traditional gradient-based learning algorithms which have to calculate the error (the difference of target $\bar{y}(t)$ and output $y(t)$) first. It is a question of why BRP works. Here we give an example to explain the plausible reasons intuitively.

\begin{figure}[htp]
\centering
\includegraphics[width=8.5cm]{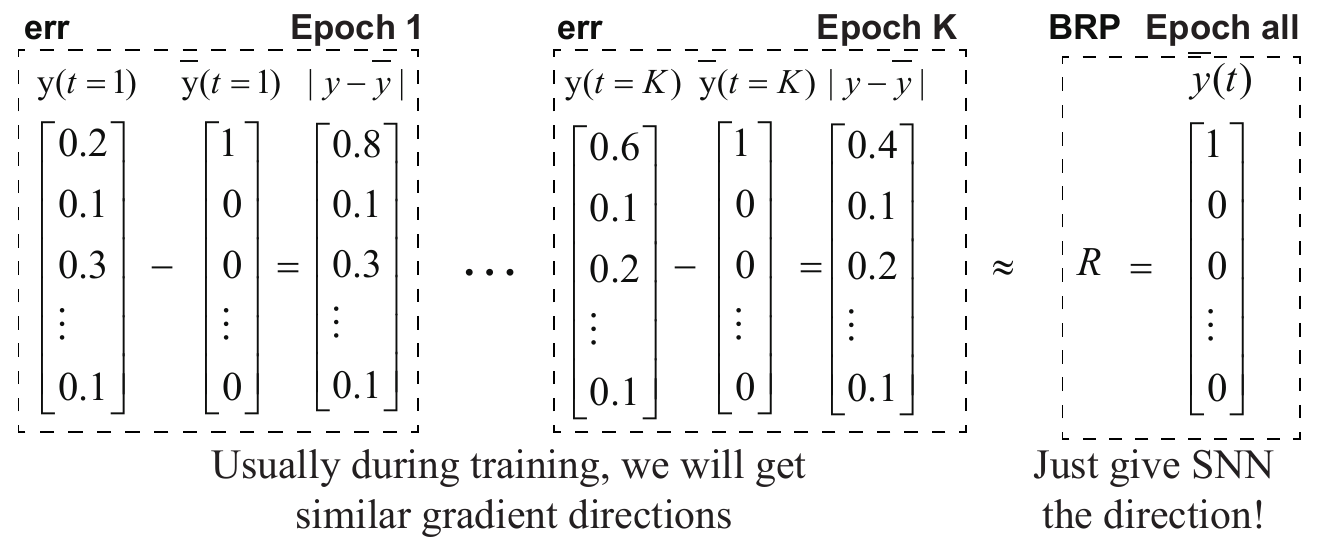}
\caption{The comparison of similar gradients between that from the BRP and that from the traditional error BP.}
\label{fig_whyworks}
\end{figure}

As shown in Fig. \ref{fig_whyworks}, the errors are usually calculated first for the next-step gradient propagation. During the training procedure, for different epochs, the synaptic weights of SNNs will be tuned with these error-related gradients. However, in the process of learning, the absolute values of errors are more dominated by the $\bar{y}(t)$ instead of the differences of $\bar{y}(t)$ and $y(t)$. Hence, given the BRP, SNN can also get a similar gradient direction. We think this is the main reason why the BRP works in the training procedure of SNNs.

\section{Experiments}\label{Experiments}

\subsection{The spatial datasets}

The MNIST \cite{RN657} and Cifar-10 \cite{RN800} datasets were selected as the two examples of spatial datasets. The MNIST dataset contains 70,000 28$\times$28 1-channel gray hand-written digit-number images from 0-9, in which 60,000 samples are training samples and the remaining 10,000 samples are testing samples. The Cifar-10 dataset contains 60,000 32$\times$32 3-channel color images covering 10 classes, 50,000 for training and 10,000 for test. The samples in both of them are static 2D images, which means a further spike generation procedure is needed.

\subsection{The temporal datasets}

The TIDigits \cite{RN798} and DvsGesture \cite{RN561} datasets were selected out as the two examples of temporal datasets. TIDigits contains 4,144 (20K HZ and around 1 second for each sample) spoken digits from 0-9. DvsGesture dataset contains 1,464 128$\times$128$\times$1800 gesture samples (low-sampled 32$\times$32 size and 100 frames from the raw DVS video with event-based 1,800 frames for the ease of computation), covering $11$ gesture types. These two datasets are challenging in temporal information processing. For example, the TIDigits dataset contains sequential pronounces of spoken numbers, and the DvsGesture dataset contains sequential videos different from clockwise and counterclockwise arm content rotations. Signals in TIDigits are continuous, which means a further procedure of spike generation is needed. Signals from the DvsGesture dataset are recorded from the DVS camera, representing natural spikes without an additional spike-conversion procedure.

\subsection{The configuration of parameters}

Parameters of the BRP-SNN for different tasks were different from each other after a little refinement to reach better performances. The refined network topologies are shown in Table \ref{tab_parameters}, including the number of spiking-convolution layers and the number of full-connection layers.

\begin{table}[htb]
\caption{BRP-SNN parameters for different tasks}
\centering
\begin{tabular}{|c|c|c|}
\hline
Tasks & Topology & Parameters \\
\hline
MNIST & Cov5*5x28-28-FC1000-FC10 & $I_{rd}(\alpha)$=1, $\eta$=1e-4 \\
\hline
Cifar-10 & Cov5*5x32-S2-FC1000-FC10 & $I_{rd}(\alpha)$=1, $\eta$=1e-4 \\
\hline
TIDigits &Cov1*3x100-S1-FC1000-FC10 & $I_{rd}(\alpha)$=0.1, $\eta$=1e-4 \\
\hline
DvsGesture & Cov5*5x32-S2-FC1000-FC10 & $I_{rd}(\alpha)$=1, $\eta$=1e-4 \\
\hline
\end{tabular}
\label{tab_parameters}
\end{table}

These parameters are not carefully designed (e.g., constructing a very deep architecture) for reaching the best performance, just staying startlines of simple shallow architectures for easier comparisons. Both $\eta^{conv}$ and $\eta^{fc}$ were set as 1e-4. For LIF neurons, $C$=~1 $uF/cm^2$, $g$=~0.2 $nS$, $\tau_{ref}$=~1$ms$, $T$=10-100 $ms$, $V_{th}$=0.5 $mV$, $V_{reset}$=0 $mV$. The local pseudo gradient was calculated with MSE loss and Adam optimizer. The batch size for all tasks was normalized as 50.

\subsection{The convergence learning of BRP-SNNs}

\begin{figure}[htp]
\centering
\includegraphics[width=8.8cm]{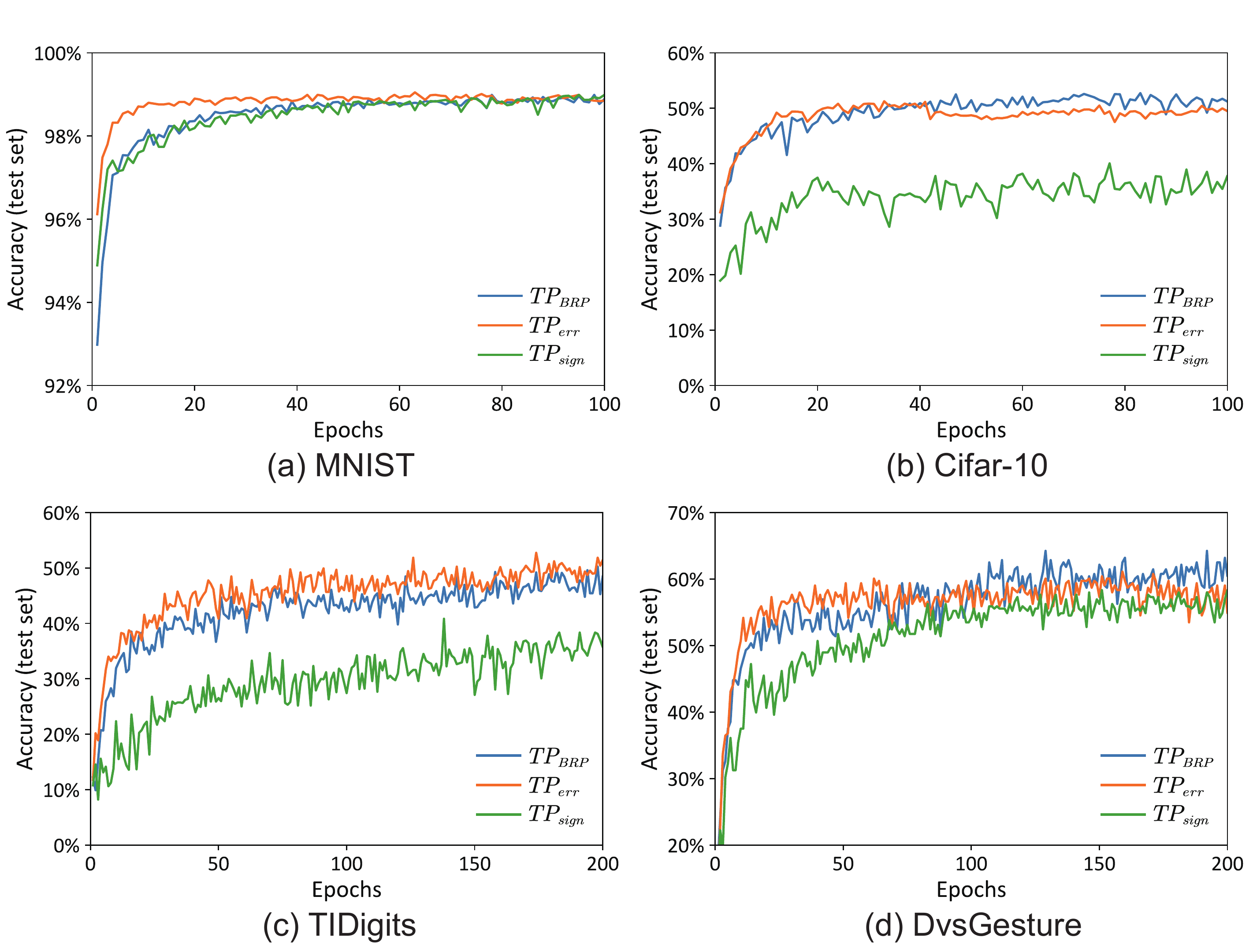}
\caption{Network convergence with different TPs on SNNs}
\label{fig_TPs_convergence}
\end{figure}

The learning convergence is usually the premise of the next-step analysis of efficient learning. As shown in Fig. \ref{fig_TPs_convergence}, three types of biologically plausible propagations were tested during the learning procedure of four datasets, including the proposed BRP ($TP_{BRP}$), the error propagation ($TP_{err}$), and sparse error sign propagation ($TP_{sign}$). The time window $T$ for these datasets was predefined as $T$=20 (for MNIST, Cifar-10, and TIDigits datasets) and $T$=30 (for DvsGesture dataset) to get better performances. The learning results on both training and test sets showed that all three biologically-plausible propagations were convergent on the four datasets.

For the MNIST dataset, the test accuracy on the SNN using $TP_{BRP}$ was convergent with the increasement of training epochs from 0 to 100. SNNs using $TP_{BRP}$ got a convergence rate higher than that using $TP_{sign}$ but lower than that using $TP_{err}$. The final performances for $TP_{BRP}$, $TP_{err}$ and $TP_{sign}$ reached 98.89\%, 98.85\% and 98.99\%, respectively.

Similar with MNIST, for the Cifar-10 and TIDigits datasets, SNNs using $TP_{BRP}$ were also convergent by reaching convergence rates higher than that using $TP_{sign}$ and comparable to that using $TP_{err}$. The final test accuracies for SNNs using $TP_{BRP}$, $TP_{err}$ and $TP_{sign}$ reached 51.21\%, 49.48\% and 37.85\% on the Cifar-10 dataset, respectively; and they reached 48.55\%, 51.45\% and 35.69\% on the TIDigits dataset, respectively.

\begin{figure}[htp]
\centering
\includegraphics[width=8.8cm]{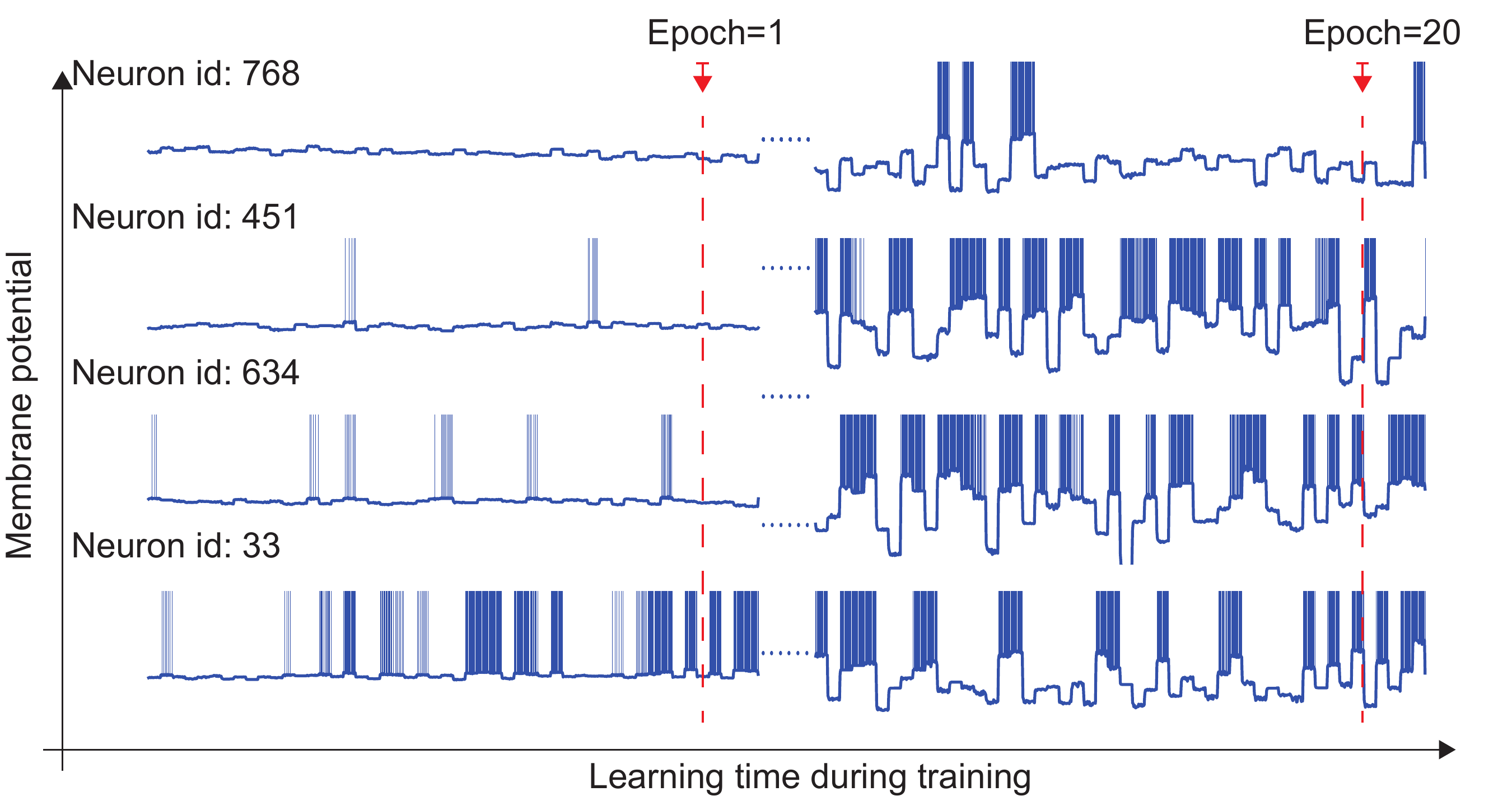}
\caption{Membrane potentials with spikes during training of the BRP-SNN}
\label{fig_mem}
\end{figure}

For the DvsGesture dataset, the $TP_{BRP}$ outperformed $TP_{err}$ and $TP_{sign}$ by reaching both higher accuracy and higher convergence rate. The final accuracies for $TP_{BRP}$, $TP_{err}$ and $TP_{sign}$ were 60.42\%, 54.86\% and 58.33\%, respectively.

Furthermore, membrane potentials with spikes were also plotted for some neurons in full-connection layers of the BRP-SNN. The indices of neurons were randomly selected, as shown in Fig. \ref{fig_mem}. The neural states were dynamically changed during the learning procedure of the BRP-SNN from epoch 1 to epoch 20, towards a better information representation (with more frequent and more stable neuron firings). In order to achieve higher performances, additional experiments were also designed with larger epochs. Finally, SNNs using BRP reached 99.01\% (for the MNIST) and 52.56\% (for the Cifar-10) after 400 epochs, and reached 65.68\% (for the TIDigits) and 64.93\% (for the DvsGesture) after 10,000 epochs. The high performance on four benchmarks showed the efficiency of the proposed BRP-SNN.

\subsection{BRP reached comparable performance with pseudo-BP}

\begin{figure}[htp]
\centering
\includegraphics[width=8.8cm]{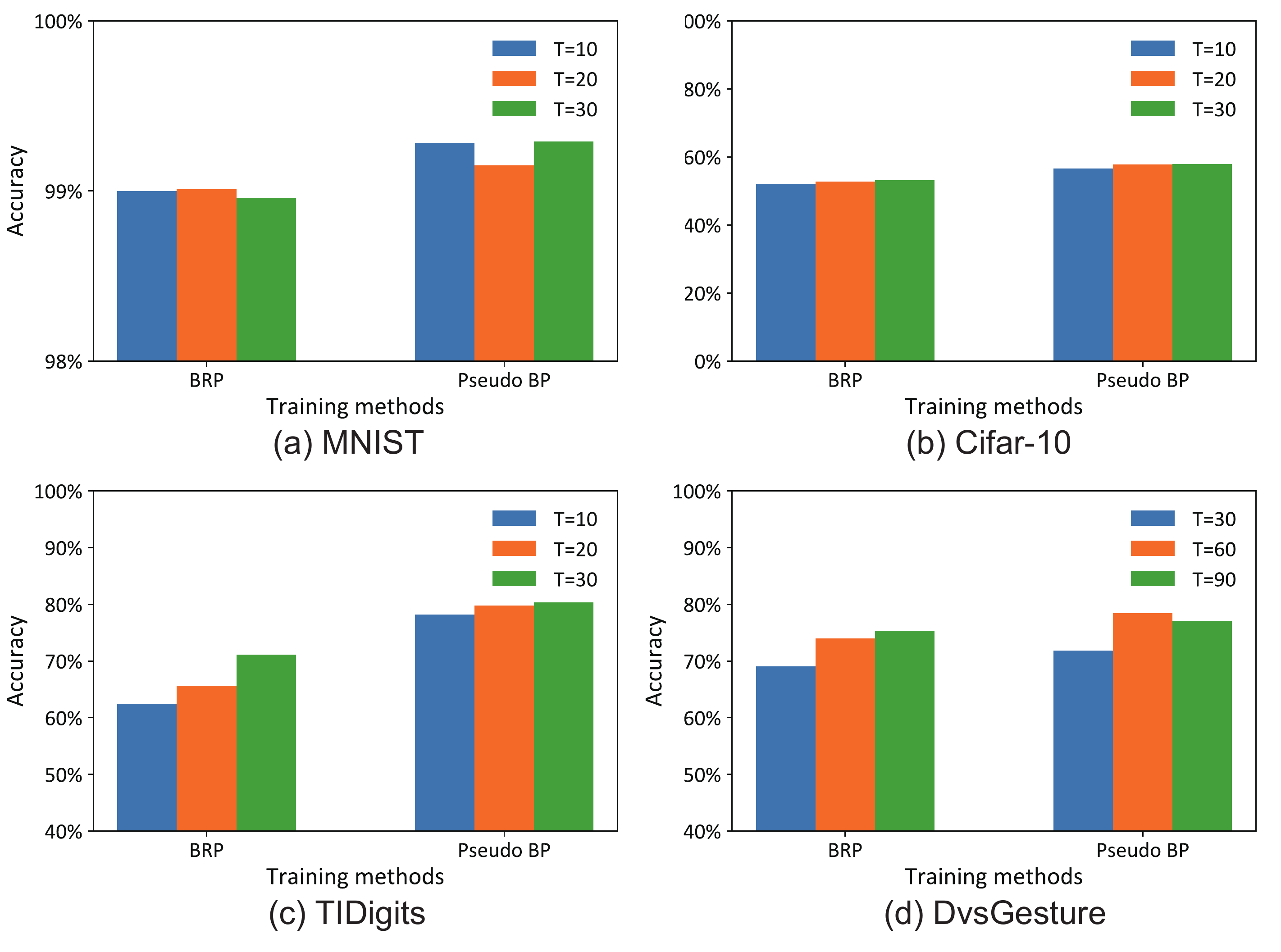}
\caption{Accuracy comparisons of spatial and temporal tasks between SNNs using BRPs and that using pseudo-BPs with different $T$.}
\label{fig_BRP_BP_compare}
\end{figure}

The inner-neuron time window $T$ described the encoding ability of neuronal dynamics, especially for temporal information. As shown in Fig. \ref{fig_BRP_BP_compare}(a,b), for spatial datasets, accuracies of SNNs trained with the BRP for different $T$ (from 10 to 30) were similar (around 98.99$\pm$0.01\% for the MNIST and around 52.65$\pm$0.61\% for the Cifar-10). One of the main reasons for this result was that the spatial data did not contain any temporal information because the temporal spike trains used for classification were usually generated from the random spike generator. The naturally-temporal datasets further verified this hypothesis by SNNs using different $T$ in Fig. \ref{fig_BRP_BP_compare}(c,d), where the accuracies of SNNs on temporal tasks (from 62.46\% to 71.14\% on the TIDigits, from 69.10\% to 76.04\% on the DvsGesture) showed a linearly-increase relationship with the size of time window $T$ (from 10 to 30 on the TIDigits and 30 to 90 on the DvsGesture).

Furthermore, the performance of SNNs using BRPs is similar to that using standard pseudo-BPs. The SNNs using BRPs and pseudo-BPs, respectively, reached accuracies of (99.01\% and 99.29\%) on the MNIST, (53.11\% and 57.88\%) on the Cifar-10, (71.14\% and 80.31\% with 1D kernels, 94.86\% and 95.10\% with 2D kernels) on the TIDigits, (76.04\% and 78.21\%) on the DvsGesture, respectively. In conclusion, the performance differences were no more than 0.28\%, 4.77\%, 0.24\%, and 2.17\%, on the MNIST, the Cifar-10, the TIDigits, and the DvsGesture datasets, respectively.

\subsection{The comparison of BRP-SNNs with other SOTA algorithms}

We further tested the performance of our BRP-SNN with other SOTA algorithms, including spike-based and rate-based SNNs. Some non-biologically-plausible algorithms were also listed for a better comparison. As shown in Table \ref{tab_SOTA}, for the MNIST task, the best accuracy of SNNs was trained with BP, where signals were represented with both firerates and spikes, and contained convolution and recurrent loops (99.6\%). Our BRP-SNN algorithm reached the accuracy of 99.01\%, higher than other SNNs tuned with STDPs.

\begin{table*}[htb]
\caption{The performance comparisons of the proposed BRP-SNN with other SOTA algorithms on the four datasets}
\centering
\begin{tabular}{|c|c|c|c|c|}
\hline
Task & Architecture & Training Type & Learning Rule & Performance \\
\hline
\multirow{4}{*}{MNIST} & Three-layer SNN~\cite{RN761} & Spike-based & Equilibrium learning + STDP & 98.52\% \\
& Three-layer SNN~\cite{RN760} & Spike-based & Balanced tuning +STDP & 98.64\% \\
& Convolutional-Recurrent SNN~\cite{lisnn} & Spike-based & Pseudo BP & 99.5\% \\
& Convolutional SNN~\cite{diehl2015fast} & Rate-based & Backpropagation & 99.1\% \\
& Recurrent SNN~\cite{RN895} & Spike-Rate-based & Pseudo BP & 99.6\% \\
& \textbf{2D-Convolutional SNN (Ours)} & \textbf{Spike-based} & \textbf{BRP} & \textbf{99.01\%} \\
\hline
\multirow{2}{*}{Cifar-10} & Three-layer SNN\cite{RN762} & Spike-based & Curiosity+STDP & 52.85\% \\
&Two-convolutional spiking layers \cite{RN643} & Spike-based & Spatio-temporal BP & 50.70\% \\
& Deep(9-11 layers) SNN \cite{lee2020fronti} & Spike-based & Direct ANN-to-SNN conversion & 91.55\% \\
& \textbf{2D-Sampling-Conv SNN (Ours)} & \textbf{Spike-based} & \textbf{BRP} & \textbf{53.11\%} \\
& \textbf{2D-Convolutional SNN (Ours)} & \textbf{Spike-based} & \textbf{BRP} & \textbf{57.08\%} \\
\hline
\multirow{4}{*}{TIDigits} & SOM SNN~\cite{wu2018a} & Spike-Rate-based & SOM+BP & 97.40\% \\
& Liquid State Machine~\cite{zhang2015a} & Spike-based &BP & 92.30\% \\
& \textbf{1D-Sampling-Conv SNN (Ours)} & \textbf{Spike-based} & \textbf{BRP} & \textbf{71.14\%} \\
& \textbf{1D-Convolutional SNN (Ours)} & \textbf{Spike-based} & \textbf{BRP} & \textbf{85.85\%} \\
& \textbf{2D-Convolutional SNN (Ours)} & \textbf{Spike-based} & \textbf{BRP} & \textbf{94.86\%} \\
\hline
DvsGesture & Deep convolution-recurrent SNN~\cite{xing2020fronti} & Spike-based &BPTT & 90.28\% \\
& \textbf{2D-Sampling-Conv SNN (Ours)} & \textbf{Spike-based} & \textbf{BRP} & \textbf{76.04\%} \\
& \textbf{2D-Convolutional SNN (Ours)} & \textbf{Spike-based} & \textbf{BRP} & \textbf{80.90\%} \\

\hline
\end{tabular}
\label{tab_SOTA}
\end{table*}

For the Cifar-10 dataset, we only made a further comparison with \cite{RN762} for its biologically plausible learning principles (with curiosity and the STDP). Our BRP-SNN reached 53.11\% accuracy, which is higher than the curiosity-based SNN (52.85\%).

For the TIDigits dataset, we got 71.14\% and 94.86\% accuracies on 1D and 2D convolutional SNNs, respectively. For the liquid state machine SNN, the accuracy only reached 92.30\%. For the integration of ANNs (e.g., Self-Organizing Map, SOM) and SNNs, the accuracy was higher than ours and reached 97.40\%. This effort showed the power of the hybrid architecture by the integration of both SNNs and ANNs.

For the DvsGesture dataset, our 2D-convolutional SNN reached 76.04\% accuracy with random sampling and 80.90\% without that, which, as far as we know, was the first time to use a biologically plausible learning principle to train spike-based SNNs without any preprocessing and cooperation with other methods. These results showed the efficiency of the proposed BRP-SNN.

\subsection{Low computation cost of the BRP-SNN with silent neurons}

In ANNs, some signals with minimal values are surprisingly important. For example, the BP gradient values that smaller than $1e-6$ (still play roles on synaptic modifications), and the neuron with small output firerates but still relatively higher than that of other output neurons as the target output class. Hence, ANNs have to take more computational resources to deal with these special situations.

In SNNs, the basic unit of signals is a discontinuous spike. The neurons that have not received any spikes or received but still not reached the firing threshold could be considered silent neurons. These silent neurons will take nearly no computational cost on specially designed neuromorphic chips \cite{davies2018loihi}. Hence, the number of these silent neurons in different layers of SNNs during network learning can be a good indicator for the computational cost, combined with the network's algorithm complexity for reaching the same accuracy.

\begin{figure}[htp]
\centering
\includegraphics[width=8cm]{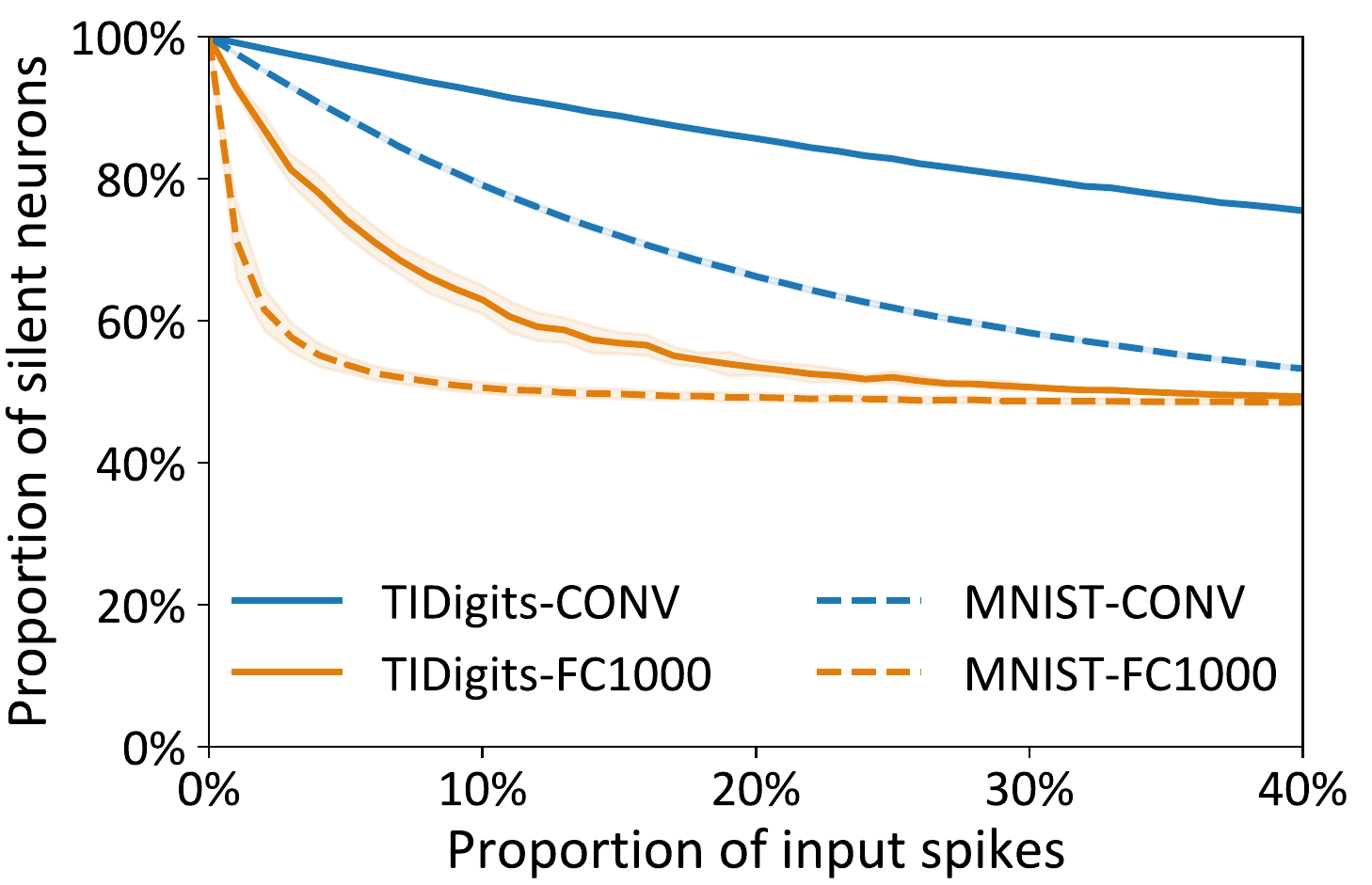}
\caption{The proportions of silent neurons in different hidden layers of SNNs (trained on the MNIST and the TIDigits datasets), with proportional increasements of input spikes in the first layer.}
\label{fig_cost}
\end{figure}

As shown in Fig. \ref{fig_cost}, the MNIST and TIDigits datasets were selected out as two representatives of spatial and temporal tasks, respectively. The proportion values of silent neurons were all decreased with the proportional increment of input spikes (from 0\% to 40\%) in different hidden layers of SNNs trained with the MNIST and the TIDigits datasets. More than half of the neurons were silent for spiking-convolution (CONV) and full-connection (FC1000 with 1000 hidden neurons) layers. These proportion values were reduced when a more significant proportion value of input spikes was given. For example, when the proportion of input spikes reached 40\%, the proportion values of silent neurons reached 53.26\% (CONV) and 48.52\% (FC1000) on the spatial MNIST dataset, which were much smaller (means the less computational cost) than that on the temporal TIDigits dataset, reached 75.50\% (CONV) and 49.32\% (FC1000).

Hence, SNNs, to this extent, at least saved more than 50\% computational cost, on the perspective of a relatively much smaller number of computing neurons (more silent neurons) than in ANNs. Besides, a lower computational cost might be more likely on temporal tasks than on spatial tasks.

\section{Conclusion}\label{Conclusion}

During evolution, the biological brain is increasingly efficient in representing and processing complex signals with spikes, delicately designed network structures, and efficient biologically plausible plasticity principles. SNNs are from biological networks' inspiration, containing discontinuous spikes instead of continuous firerates for the information propagation. These seem-simple spikes are not trivial but have played essential roles in the computation of sparse (or event-based) signals. They also reflect the complex inner computation of neuronal dynamics and the balance between higher accuracy and lower computational cost. SNNs also contain various types of plasticity principles, including local plasticity principles (e.g., STDP, STP, Hebb, lateral inhibition) and global plasticity principles (e.g., dopamine-like reward that propagates directly along with top-down connections from reward neurons to target neurons, without any multi-step routing).

In this paper, we focus on both spikes from neuronal dynamics and biologically-plausible learning principles for the good tuning of SNNs, not for the highest accuracy, but an alternative effort to understand a possibly efficient tuning strategy better in the real biological brain. Inspired by the brain's direct reward signal propagation, we have designed the SNN using Biologically-plausible Reward Propagation (BRP-SNN) to directly propagate the right label index (with the form of label-spike trains) to target neurons. Spiking-convolution and full-connection layers with dynamic LIF neurons have been constructed for the BRP-SNN and verified on four benchmark datasets, including spatial datasets (the MNIST and the Cifar-10) and also temporal datasets (the TIDigits and the DvsGesture). The BRP-SNN has got comparable performances compared to that tuned with SOTA pseudo-BPs under the same SNN architectures, reaching testing accuracy with 99.01\% for the MNIST, 57.08\% for the Cifar-10, 94.86\% for the TIDigits, and 80.90\% for the DvsGesture, respectively. Besides, a further computation-efficiency analysis has been given, showing that the BRP-SNN has saved at least 50\% computational cost on both full-connection layers and spiking-convolution layers compared to that in standard ANNs.

The standard BP in ANNs back propagates error signals from the final output neurons to previous neurons layer by layer, which is not biologically plausible and not energy-efficient. We think the deeper research on biologically plausible plasticity principles (e.g., BRP or other local plasticity principles such as STDP and STP) will someday finally replace the BP, toward human-level robust computation, energy-efficient learning, or even artificial general intelligence that the community of artificial intelligence is facing but still has not been resolved. We think this research will also give us more hints on a better understanding of the biological system's intelligent nature and promote a better study of globally neural plasticity in biological networks.

\section*{Acknowledgment}
We thank Ruichen Zuo and Kaiyuan Liu for their assistance with the final version paper checking. This study is supported by the National Key R\&D Program of China (Grant No. 2020AAA0104305), the National Natural Science Foundation of China (Grant No. 61806195), the Strategic Priority Research Program of the Chinese Academy of Sciences (Grant No. XDB32070100, XDA27010404), and the Beijing Brain Science Project (Z181100001518006). The source code of BRP-SNN is in https://github.com/thomasaimondy/BRP-SNN.

\ifCLASSOPTIONcaptionsoff
\newpage
\fi

\bibliographystyle{IEEEtran}
\bibliography{thomas}

\newpage
\vspace{-80 mm}
\begin{IEEEbiography}[{\includegraphics[width=1in,height=1.0in,clip,keepaspectratio]{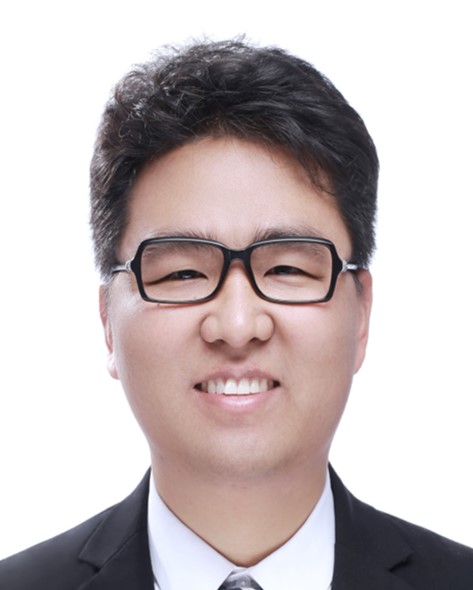}}]{Tielin Zhang}
received the Ph.D. degree from the Institute of Automation Chinese Academy of Sciences, Beijing, China, in 2016. He is an Associate Professor in the Research Center for Brain-Inspired Intelligence, Institute of Automation, Chinese Academy of Sciences, Beijing, China. His current interests include theoretical research on neural dynamics and Spiking Neural Networks (more information is in https://bii.ia.ac.cn/$\sim$tielin.zhang/).
\end{IEEEbiography}

\vspace{-80 mm}
\begin{IEEEbiography}[{\includegraphics[width=1in,height=1.0in,clip,keepaspectratio]{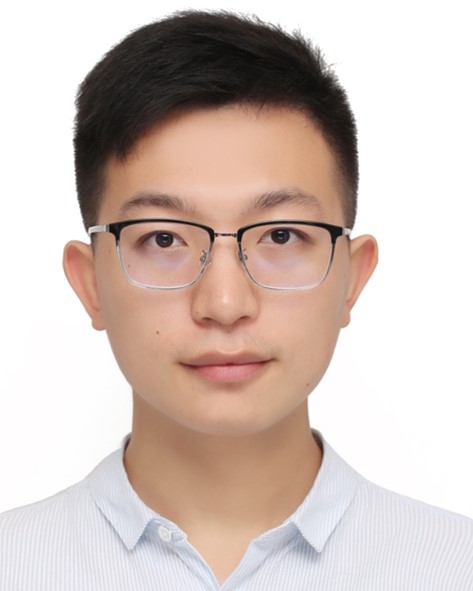}}]{Shuncheng Jia}
is a Ph.D. candidate in both the Institute of Automation Chinese Academy of Sciences and the University of Chinese Academy of Sciences. His current interests include theoretical research on neural dynamics, auditory signal processing, and Spiking Neural Networks.
\end{IEEEbiography}

\vspace{-80 mm}
\begin{IEEEbiography}[{\includegraphics[width=1in,height=1.0in,clip,keepaspectratio]{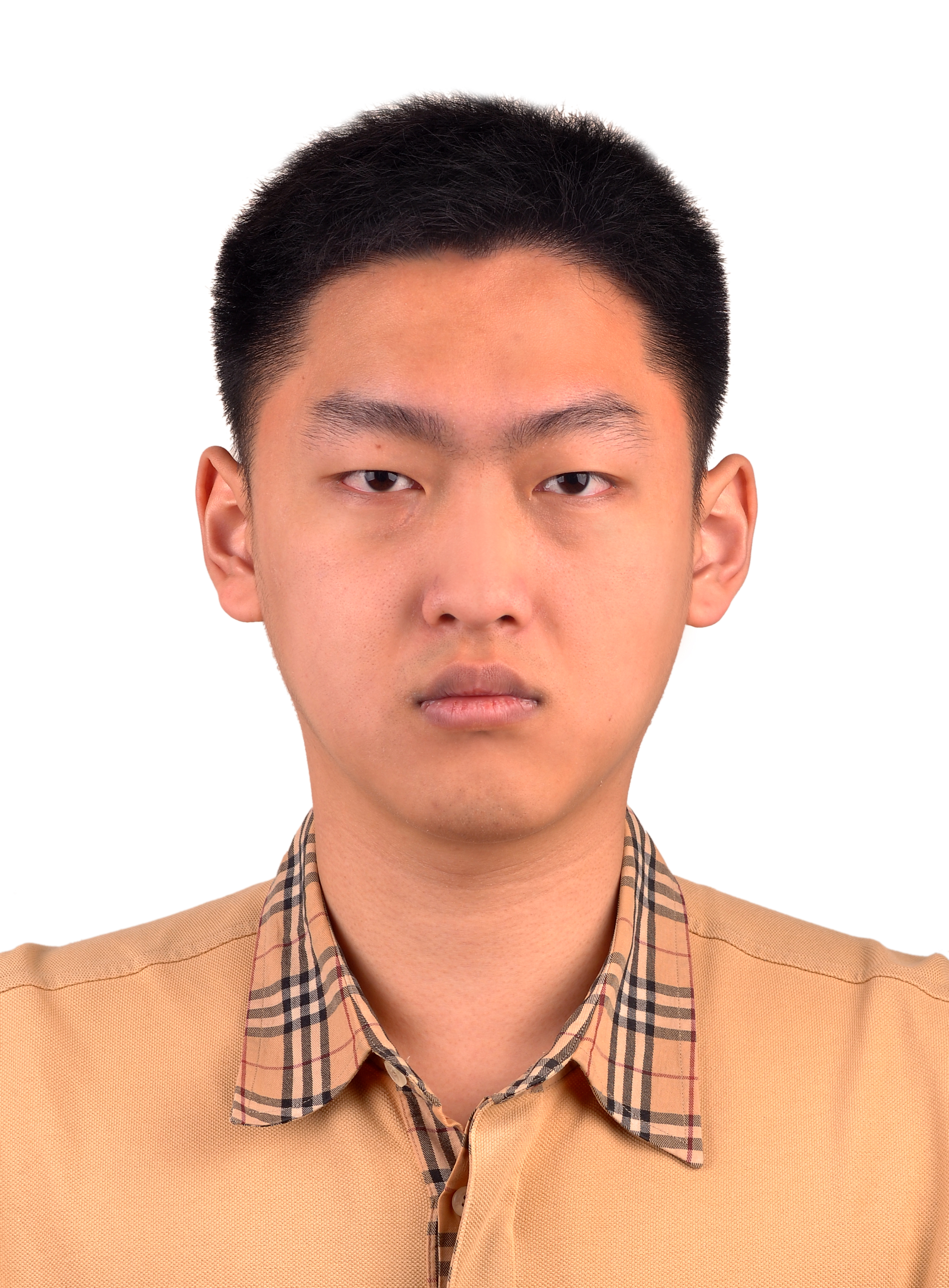}}]{Xiang Cheng}
is a Ph.D. candidate in both the Institute of Automation Chinese Academy of Sciences and the University of Chinese Academy of Sciences. His current interests include biological informatics, neural dynamics, spatio-temporal signal processing, and Spiking Neural Networks.
\end{IEEEbiography}

\vspace{-80 mm}
\begin{IEEEbiography}[{\includegraphics[width=1in,height=1.0in,clip,keepaspectratio]{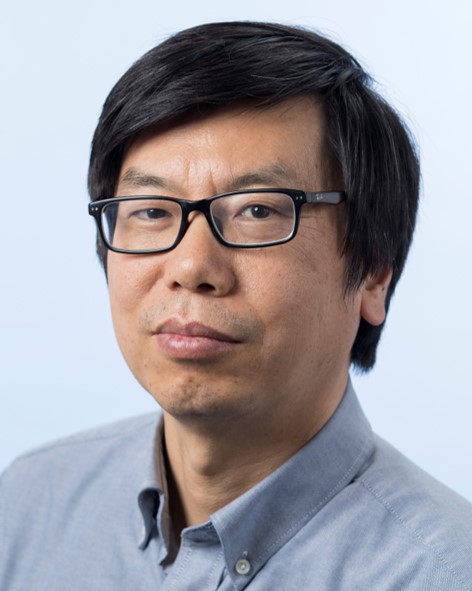}}]{Bo Xu}
is a professor, the director of the Institute of Automation Chinese Academy of Sciences, and also deputy director of the Center for Excellence in Brain Science and Intelligence Technology, Chinese Academy of Sciences. His main research interests include brain-inspired intelligence, brain-inspired cognitive models, natural language processing and understanding, brain-inspired robotics.
\end{IEEEbiography}

\end{document}